%% file: manuscript.tex
\RequirePackage{fix-cm}
\documentclass[10pt,letterpaper]{article}
\usepackage{spconf}
\usepackage[T1]{fontenc}
\usepackage{amsmath,amssymb,array,booktabs,tabularx,graphicx}
\usepackage{tikz}
\usetikzlibrary{arrows.meta,positioning,calc}
\usepackage{cite,url,microtype}
\usepackage{balance}
\makeatletter
\patchcmd{\@BAlancecol}{\setbox\@leftcolumn}{\advance\@tempdima by 2pt\setbox\@leftcolumn}{}{}
\makeatother
\usepackage[hidelinks]{hyperref}
\hypersetup{pdftitle={NAWE: Digital Watermarking with Neural-Assisted Watermark Extraction},
pdfauthor={Roman Chaban, Vitaliy Kinakh, Lilian Rouzaire, Slava Voloshynovskiy},
pdfsubject={Signal-processing watermarking with neural host suppression}}

\title{NAWE: Digital Watermarking with\\Neural-Assisted Watermark Extraction}
\name{Roman Chaban$^{\star}$, Vitaliy Kinakh$^{\star}$, Lilian Rouzaire$^{\dagger}$ and Slava Voloshynovskiy$^{\star}$}
\address{$^{\star}$University of Geneva, Switzerland \qquad $^{\dagger}$Independent Researcher\\
\{roman.chaban, vitaliy.kinakh, svolos\}@unige.ch, lilian.rouzaire@gmail.com}
\begin{document}
\ninept
\maketitle
\begin{abstract}
NAWE (Neural-Assisted Watermark Extraction) combines an explicit signal-processing watermarking construction with a pretrained neural host predictor. A periodic, perceptually masked watermark carrier provides synchronization, Polar coding supplies redundancy, and denoising followed by subtraction extracts the embedded watermark. The denoiser remains frozen, without watermark-specific training. A one-factor-at-a-time study compares Wiener, BM3D, DRUNet, and GS-DRUNet host estimators. Comparisons with TrustMark, SSL Watermarking, PixelSeal, and WAM show NAWE's lowest geometric and photometric class BER and strong message recovery, while filtering and noise remain limitations consistent with the non-adaptive selection of the watermark extractor. The comparison retains the systems' different payloads and coding.

The code will be available at GitHub on paper acceptance.
\end{abstract}

\section{Introduction}
Invisible image watermarking must preserve visual quality while recovering a message after image processing and geometric distortion. Filtering and compression corrupt watermark evidence; cropping, resizing, and rotation also disturb carrier alignment. Robustness therefore requires synchronization, encoding, and suitable watermark extraction for efficient host-interference suppression.

Classical signal-processing designs include spread-spectrum modulation \cite{cox1997secure}, perceptually masked transform-domain embedding \cite{barni1998dct}, and stochastic content-adaptive allocation and host estimation \cite{voloshynovskiy2000adaptive}. Quantization index modulation \cite{chen2001qim}, known-host data hiding \cite{perez2003knownhost}, and improved spread spectrum \cite{malvar2003iss} address host interference at embedding or detection. Geometric robustness has been pursued through transform invariants \cite{oruanaidh1998rotation}, registration templates \cite{pereira2000template}, and repeated spatial patterns with autocorrelation-based acquisition \cite{kutter1998geometry}. Subsequent periodic and self-reference constructions combine this synchronization principle with multibit coding and adaptive embedding \cite{volosh2001multibit,volosh2001diversity,deguillaume2003hybrid}.

Learned systems extend this tradition through robust representations and distortion-aware embedder/extractor training \cite{fernandez2022ssl,bui2025trustmark,sander2025wam,pixel-seal2025}. NAWE instead uses a neural model for a specific receiver task: predicting the host so that subtraction extracts the watermark, thus providing the host interference cancellation. A coded, repeated spatial watermark carries out an encrypted message and supplies synchronization, and registered repetitions support soft decoding. The denoiser is pretrained and frozen; the watermarking system requires no task-specific network training.

Our contribution integrates explicit synchronization with replaceable neural host suppression, compares conventional and neural predictors, and evaluates bit and word error rates, and attack-family behavior.

\input{figures/pipeline.tex}

\section{Relation to learned watermarking}
\textbf{SSL Watermarking} \cite{fernandez2022ssl} freezes a self-supervised trained network and optimizes each image's pixels by adversarial embedding: gradient-based updates make feature projections onto keyed directions match the signs specified by the message under a family of defined distortions. Extraction reads those signs without the host. The configurable message length $L_{\rm msg}$ counts direct bits without an outer channel code.

\textbf{TrustMark} \cite{bui2025trustmark} trains a residual embedder and a message extractor with image distortions; both operate feed-forward at deployment. The Q/BCH-5 interface uses 100 carrier bits: $L_{\rm msg}=61$ information bits, 35 BCH parity bits, and $L_{\rm schema}=4$ schema/reserved bits \cite{trustmark2024code}. Channel decoding follows neural carrier extraction.

\textbf{WAM} \cite{sander2025wam} trains an embedder and an extractor that jointly localizes marked pixels and predicts a message of length $L_{\rm msg}=32$ bits at each location. Spatial aggregation recovers distinct messages from marked regions. Training includes localized embedding, splicing, and geometric/photometric augmentations.

\textbf{PixelSeal} \cite{pixel-seal2025} trains a feed-forward embedder/extractor with message length $L_{\rm msg}=256$ bits, distortion augmentation, and high-resolution adaptation. An adversarial discriminator supplies the imperceptibility objective instead of fixed perceptual losses, alongside the message-recovery objective. The adversarial optimization occurs during training.

NAWE embeds $L_{\rm msg}$ bits on the native image grid without learnable models and uses a frozen neural denoiser only at the extraction.

\section{Periodic embedding and blind extraction}
\subsection{Message coding and spatial construction}
Figure~\ref{fig:pipeline} summarizes the signal path of NAWE. Let $\mathbf x\in[0,1]^{H\times W\times C}$ be the host and $\mathbf b_{\rm msg}\in\{0,1\}^{L_{\rm msg}}$ an image-independent message, with $16\le L_{\rm msg}\le256$ bits. Appending cyclic redundancy check (CRC) bits of length $L_{\rm CRC}=16$  gives total length $L=L_{\rm msg}+L_{\rm CRC}$; the CRC carries no user information ($L_{\rm CRC}$ may be 0); it flags residual errors and steers list decoding (Sec.~\ref{sec:acq}). The resulting word and the encoded codeword are
\begin{align}
\mathbf u&=[\mathbf b_{\rm msg}\|\operatorname{CRC}_{16}(\mathbf b_{\rm msg})]\in\{0,1\}^{L},\label{eq:frame}\\
\mathbf c&=\operatorname{PolarEnc}_{N,L}\bigl(\mathcal E_{k_p,\nu}(\mathbf u)\bigr).\label{eq:code}
\end{align}
Polar coding maps an $L$-bit input word to an $N$-bit codeword, where $N$ is a power of two. The $L$ input bits occupy positions corresponding to the most reliable polarized channels; the remaining $N-L$ positions are fixed to known values, called frozen bits \cite{arikan2009polar}. The code rate is $L/N$. For example, choosing $L=3N/4$ gives rate $3/4$ while transmitting all $N$ encoded bits, without puncturing (removing encoded bits).

The length-preserving encryption $\mathcal E_{k_p,\nu}$ uses the shared secret key $k_p$ and a nonce $\nu$, a public per-embedding value supplied to the receiver to diversify encryption.

The key/nonce pair $(k_p,\nu)$ also generates the known pseudorandom synchronization vector $\mathbf s\in\{0,1\}^{L_\mathbf{s}}$ of length $L_\mathbf s$. These additional bits occupy dedicated cells alongside $\mathbf c$; they are not user information or Polar parity bits. The receiver regenerates $\mathbf s$ to resolve orientation and translation, refine the affine estimate, and validate alignment (see Sec.~\ref{sec:acq}). A shared layout maps the two bit sequences to a two-dimensional array, then bipolar modulation maps $0$ to $+1$ and $1$ to $-1$:
\begin{equation}
\mathbf w_{\rm blk}=1-2\operatorname{Layout}(\mathbf c,\mathbf s).
\label{eq:layout}
\end{equation}
Expanding cells by integer scale $s$ (distinct from $\mathbf s$) and repeating the block produces the carrier $\mathbf w$ on the native image grid. Repetitions provide additional redundancy for the same word. The construction leaves $N$, $L_{\rm msg}$, $L_{\rm CRC}$, $L_\mathbf s$, and $s$ configurable.

\begin{figure}[!t]
\vspace{-3mm}
\centering\includegraphics[width=.86\columnwidth]{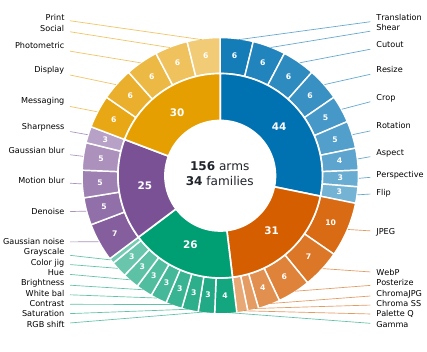}
\vspace{-6mm}
\caption{Composition of the 156-arm benchmark across 34 attack families. Inner sectors: geometry (44), photometric/color (26), compression/quantization (31), filtering/restoration/noise (25), and compound (30). Outer sectors show severity-arm counts.}
\label{fig:benchmark}
\vspace{-2mm}
\end{figure}

\subsection{Perceptual allocation and host suppression}

The host determines a perceptual mask that is smaller in flat
regions and larger in textured regions [3]:
\begin{equation}
\mathcal{M}(\mathbf{x})
=\frac{\sigma^{2}+\gamma\beta^{2}}{\sigma^{2}+\beta^{2}},
\qquad
\mathbf{m}
=\frac{\mathcal{M}(\mathbf{x})}
{\operatorname{RMS}\!\left(\mathcal{M}(\mathbf{x})\right)}.
\label{eq:perceptual_mask}
\end{equation}
Here, $\sigma^{2}$ is the local image variance and
$\gamma\in(0,1]$ is the flat-region weight before normalization.
We set $\beta=40$ when the variance is computed on the 8-bit
scale, equivalently $\beta=40/255$ on the $[0,1]$ scale.
The mask $\mathbf{m}$ has unit root mean square (RMS), while $\alpha$ controls
the overall embedding strength. Embedding is
\begin{equation}
\mathbf y=Q\!\left(\operatorname{clip}_{[0,1]}[\mathbf x+\alpha\mathbf m\odot\mathbf w]\right).
\label{eq:embed}
\end{equation}
Here $Q$ denotes image quantization. The scalar $\alpha$ targets the embedding distortion of the saved image, with
$\rm{PSNR}(\mathbf x, \mathbf y)=-10\log_{10}(\|\mathbf x-\mathbf y\|_F^2/HWC)$.

For a received image $\mathbf z$, the host estimate and watermark residual are
\begin{equation}
\widehat{\mathbf x}=D_{\theta_0,\sigma}(\mathbf z),\qquad
\widehat{\mathbf w}=\mathbf z-\widehat{\mathbf x}.
\label{eq:residual}
\end{equation}
The denoiser has frozen weights $\theta_0$ and noise-level setting $\sigma$. GS-DRUNet \cite{hurault2022gradient} is compared with DRUNet \cite{zhang2021dpir}, BM3D \cite{dabov2007bm3d}, and Wiener filtering. Predictors are evaluated by downstream recovery.

\subsection{Autocorrelation acquisition, aggregation, and decoding}\label{sec:acq}
The autocorrelation function (ACF) measures the similarity of the residual to its spatially shifted copies. It is computed by the Wiener--Khinchin relation
\begin{equation}
R_{\widehat{\mathbf w}}=\mathcal F_2^{-1}\!\left(\left|\mathcal F_2\{\widehat{\mathbf w}\}\right|^2\right).
\label{eq:acf}
\end{equation}
Here $\mathcal F_2$ is the two-dimensional Fourier transform. The repeated carrier creates off-origin ACF peaks, following classical self-reference synchronization \cite{kutter1998geometry,volosh2001multibit,deguillaume2003hybrid}. Under $\mathbf r\mapsto\mathbf A\mathbf r+\mathbf t$, their displacements depend on the four entries of $\mathbf A\in\mathbb R^{2\times2}$, but not on the two translation components $\mathbf t$. RANSAC fits candidate lattice bases and hence affine-matrix hypotheses. Inverse warping compensates the linear transform; folding and averaging registered tiles reinforces their common watermark.

The remaining orientation and translation are resolved by cross-correlating the folded residual with the known bipolar synchronization cells $1-2\mathbf s$ generated from the key/nonce pair $(k_p,\nu)$. Candidate cyclic shifts are ranked by this correlation, yielding translation in registered coordinates modulo the tile period. Correlation with the known synchronization cells also supports local refinement and validation of the affine hypotheses. The orientation/phase candidates are then decoded.

For candidate $j$, let $\boldsymbol\ell_j$ be the soft coded symbols after removal of synchronization cells. The receiver decodes and decrypts:
\begin{equation}
\widehat{\mathbf u}_j=\mathcal E^{-1}_{k_p,\nu}
\bigl(\operatorname{PolarDec}_{N,L}(\boldsymbol\ell_j)\bigr).
\label{eq:decode}
\end{equation}
$\operatorname{PolarDec}$ is CRC-aided successive-cancellation list decoding, selecting the listed path whose CRC checks \cite{tal2015list}. Writing $\widehat{\mathbf u}_j=[\widehat{\mathbf b}_{{\rm msg},j}\|\widehat{\mathbf b}_{{\rm crc},j}]$, decoding stops and returns $\widehat{\mathbf b}_{{\rm msg},j}$ when the recomputed and decoded CRC bits agree:

\begin{equation}
\operatorname{CRC}_{16}(\widehat{\mathbf b}_{{\rm msg},j})
=\widehat{\mathbf b}_{{\rm crc},j}.
\label{eq:crc}
\end{equation}

\begin{figure}[!t]
\vspace{-10mm}
\centering\includegraphics[width=0.47\textwidth]{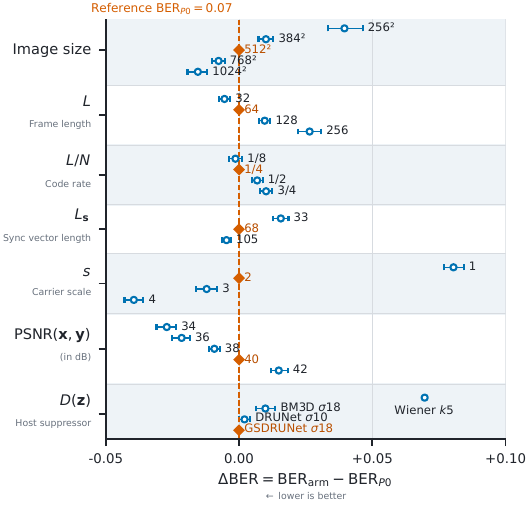}
\vspace{-4mm}
\caption{OFAT sensitivity as $\Delta\mathrm{BER}=\mathrm{BER}_{\rm arm}-\mathrm{BER}_{P_0}$ (smaller is better). $\mathrm{BER}_{\rm arm}$ is the BER of a construction with a single control parameter changed and $\mathrm{BER}_{P_0}$ that of the reference $P_0$ under the set of attacks that is comprised of 34 arms from 7 families (reduced form of Figure~\ref{fig:benchmark}), among which are JPEG, Gaussian blur and noise, resize, rotation, crop and flip on 100 images. $\mathrm{BER}_{P_0}=0.07$. Only one control parameter changes per row.}
\label{fig:ofat}
\vspace{-2mm}
\end{figure}

Candidates are tested until CRC agreement or budget $J$ is exhausted. Blind extraction uses the key, nonce, layout, and decoder settings. It targets zero-bit-error recovery of $\mathbf u$: every bit must match the transmitted word. At least one wrong bit or a decoding failure constitutes a word error. The word error rate (WER) is the fraction of trials with such an error, as defined in Sec.~\ref{sec:protocol}.

\vspace{-2mm}
\section{Evaluation and results}

\subsection{Comparison protocol and error rates}\label{sec:protocol}
We compare native implementations of competing watermarking methods on identical images and attacks at matched embedding PSNR, retaining their payloads and available error correction. These are complete-system comparisons: equal PSNR does not equalize information rate, coding gain, or perceptual visibility. Images are COCO~2017 \cite{lin2014coco} resized to $512^2$: 100 sources for the robustness benchmark and 100 for one-factor-at-a-time (OFAT) study. Each source draws an independent, domain-separated key and nonce, so key, nonce, and message never repeat.

\begin{figure}[!t]
\vspace{-10mm}
\centering\includegraphics[width=\columnwidth]{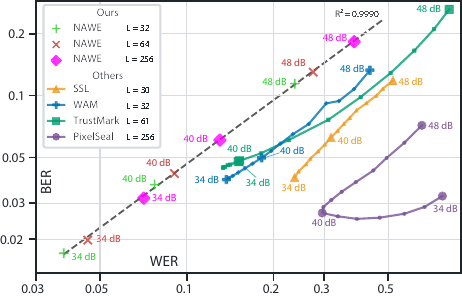}
\vspace{-7mm}
\caption{Family-balanced WER versus BER over the robustness matrix in Figure~\ref{fig:benchmark}, on logarithmic axes. Baseline curves sweep $\rm{PSNR}(\mathbf x, \mathbf y)\in[34,48]$~dB; NAWE markers show 48, 40, and 34~dB for protected-frame lengths $L\in\{32,64,256\}$, all at code rate $1/4$. Legend subscripts give the evaluated bit length.}
\vspace{-2mm}
\label{fig:tradeoff}
\end{figure}

For method $q$ on arm $a$ of attack family $f$, let $\mathbf b_i^{(q)}$ be a scored word of length $L_q$ and $\widehat{\mathbf b}_i^{(q,f,a)}$ its decoded output. With $\mathcal I_{q,f,a}$ the set of returned words we define the bit error rate (BER) and word error rate (WER) as:
\begin{align}
\operatorname{BER}_{q,f,a}&=\frac{\sum_{i\in\mathcal I_{q,f,a}}d_H(\widehat{\mathbf b}_i^{(q,f,a)},\mathbf b_i^{(q)})}
{|\mathcal I_{q,f,a}|L_q},\label{eq:ber}\\
\operatorname{WER}_{q,f,a}&=1-\frac{1}{T_{f,a}}\sum_{i\in\mathcal I_{q,f,a}}
\mathbf1\{\widehat{\mathbf b}_i^{(q,f,a)}=\mathbf b_i^{(q)}\}.\label{eq:wer}
\end{align}
Here $T_{f,a}$ counts all trials and $d_H$ is Hamming distance. BER is the fraction of wrong bits among returned words, a rate normalized by payload length, and is therefore comparable across different $L_q$. WER is the fraction of trials with at least one wrong bit or no decoded word. The indicator in \eqref{eq:wer} is one only when all $L_q$ decoded bits match the original word. For NAWE, $L_q=L=L_{\rm msg}+L_{\rm CRC}$ includes message and CRC bits after decoding.

\begin{figure*}[!t]
\vspace{-10mm}
\centering\includegraphics[width=\textwidth]{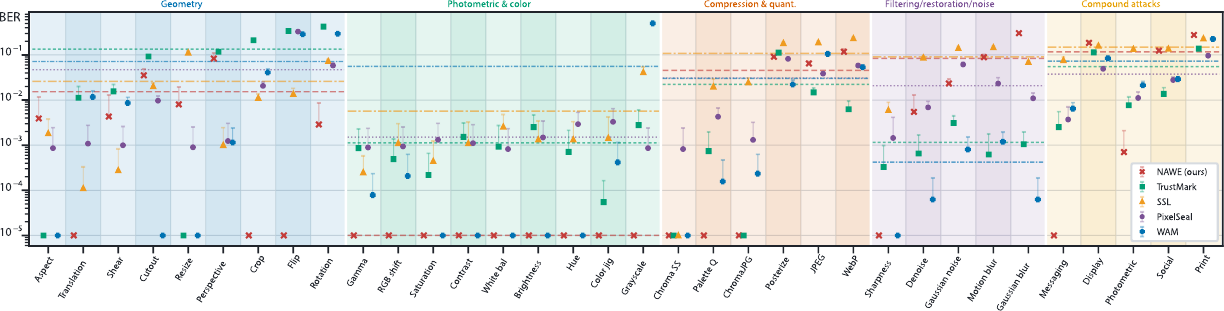}
\vspace{-7mm}
\caption{Per method BER at 40 dB across the 34 attack families from Figure~\ref{fig:benchmark} benchmark. Each marker is $\mu$ BER across all arms of a single attack from X axis and upward whisker is $\sigma$. Dashed horizontal lines represent $\mu$ of each distortion class. NAWE has the lowest geometric and photometric class averages.}
\vspace{-2mm}
\label{fig:attacks}
\end{figure*}

\subsection{Parameter sensitivity and the host predictor}
Figure~\ref{fig:ofat} varies one control around the orange-diamond \emph{reference} $P_0$: $L_{\rm msg}=48$, $L_{\rm CRC}=16$ and $L=L_{\rm msg}+L_{\rm CRC}=64$ is the Polar input length, which code rate $L/N=1/4$, $L_\mathbf s=68$ synchronization bits that occupy separate carrier cells and are excluded from Polar coding. Cell scale $s=2$, GS-DRUNet at $\sigma=18$, $\rm{PSNR}(\mathbf x, \mathbf y)=40$~dB, and RGB image is $512^2$ pixels. The $L_\mathbf s$ keyed sync bits The panel reports $\Delta\mathrm{BER}=\mathrm{BER}_{\rm arm}-\mathrm{BER}_{P_0}$ (smaller is better) against the reference $\mathrm{BER}_{P_0}=0.07$, the operating point carried into the robustness benchmark (Figs.~\ref{fig:benchmark}--\ref{fig:attacks}).

Two controls drive the largest reductions below $P_0$: the carrier scale $s$ and the host suppressor $D_{\theta_0,\sigma}$. A coarser cell ($s{=}4$) and the neural denoiser push $\Delta\mathrm{BER}$ most negative, while $s{=}1$ and $5\times5$ Wiener are the worst arms ($\Delta\mathrm{BER}\approx+0.08$ and $+0.07$); GS-DRUNet, DRUNet ($\sigma{=}10$), and BM3D ($\sigma{=}18$) lie within $\approx0.01$ of one another. The suppressor, the code rate $L/N{=}1/4$, and the sync length $L_\mathbf s{=}68$ are deliberate design choices near their best settings; too few sync cells ($L_\mathbf s{=}33$) or a higher rate cost accuracy at a redundancy trade we set once.

Embedding strength, image size, and payload are channel choices rather than design defaults, and they follow the expected monotone trend. At fixed code and cell geometry the averaged per-symbol SNR scales as $\mathrm{SNR}\propto\alpha^2\,HWC/L_{\rm msg}$, so stronger embedding (larger $\alpha$, i.e.\ lower $\rm{PSNR}(\mathbf x, \mathbf y)$), a larger host ($HWC$, more carrier repetitions), or a shorter payload each lower BER, roughly linearly across the swept range. The carrier is expanded by plain repetition on the native grid, so these operating points need no retraining, unlike the learned baselines, which require re-optimization for a new resolution or payload.

\subsection{Benchmark coverage and distortion trade-off}
Figure~\ref{fig:benchmark} groups 156 severity settings (arms) into 34 operation families $\mathcal F$ and five broad classes. Let $S_{q,f,a}$ denote the selected endpoint (BER or WER) for method $q$ on arm $a$ of family $f$, whose arm set is $\mathcal A_f$. We weight families equally:
\begin{equation}
\overline S_q=\frac{1}{|\mathcal F|}\sum_{f\in\mathcal F}\frac{1}{|\mathcal A_f|}\sum_{a\in\mathcal A_f}S_{q,f,a}.
\label{eq:aggregate}
\end{equation}
Each family contributes $1/|\mathcal F|$ regardless of its arm count, and its arms share that weight uniformly; the number of arms in a family therefore does not affect the aggregate. Class summaries apply the same rule to the families within a class. WAVES also emphasizes attack diversity \cite{an2024waves}; geometric severity requires parameters beyond pixel-aligned PSNR.

Figure~\ref{fig:tradeoff} plots WER against BER to evaluate the relation between zero-bit and multi-bit performance of NAWE and its competitors. Changing NAWE's frame length $L$ shifts its operating points along nearly linear log-log trend as expected when more coded symbols share a fixed distortion budget, without changing the construction or retraining a model. Across 34, 40, and 48~dB, NAWE has lower WER than every payload-matched baseline and competitive BER. At $\rm{PSNR}(\mathbf x, \mathbf y)<40$~db NAWE starts to outperform other baselines in both WER and BER at competitive frame length $L$. The most pronounced difference can be observed at $\rm{PSNR}(\mathbf x, \mathbf y)=34$~db, where the competitors are saturated and cannot improve beyond theirs operational boundaries.

\subsection{Attack-wise behavior and limits}
Figure~\ref{fig:attacks} compares the BER of NAWE and its competitive baselines using the benchmark composition in Figure~\ref{fig:benchmark}.

\textbf{Geometry:} NAWE has the lowest class average, explained by the explicit synchronization mechanism. Flip and translation lie at the plotting floor, and rotation, shear, and resizing have low BER. SSL is stronger on some families, notably perspective: a general projective warp exceeds the acquisition model, which in a way can be tuned for perspective transforms as well.

\textbf{Photometric and color:} NAWE attains the lowest class summary and stays at the floor throughout this group, with ties on individual operations. Grayscale is particularly challenging for WAM. Altogether the learned baselines demonstrate inconcistency under somewhat trivial and not greatly destructive distortions.

\textbf{Compression and quantization:} NAWE remains competitive, especially for chroma JPEG, chroma subsampling, and palette quantization. Standard JPEG, posterization, and WebP produce more errors, and the other systems can achieve lower BER.

\textbf{Filtering, restoration, and noise:} NAWE underperforms the competing methods in this class, particularly TrustMark and WAM. Its off-the-shelf denoiser was pretrained for additive white Gaussian noise. Highly destructive attacks that primarily target high frequencies, like denoise and blurs, are excellent watermark suppressors. Nevertheless, severe blur removes commercially useful detail and resulting attacked image has comparatively low SSIM and PSNR. This training-task mismatch limits the current design; training or adapting the host predictor for extraction under these distortions is a future research direction.

\textbf{Compound:} The compound arms chain several operations to emulate real distribution pipelines, such as messaging and social re-encoding, display capture, and print. NAWE's behavior on them tracks its per-class results: it stays robust on chains dominated by geometric and photometric steps, but degrades where a filtering or noise stage is present, inheriting the limitation above.

\section{Conclusion}
NAWE couples an explicit signal-processing construction and a periodic, perceptually masked carrier with autocorrelation-based synchronization and Polar coding -- to a frozen neural denoiser used only for host suppression at extraction. This design keeps the receiver blind and free of watermark-specific training while a neural predictor still removes host interference. Across a broad multi-family benchmark NAWE is competitive overall and leads the geometric and photometric classes, and it attains the most reliable whole-word recovery among the compared systems, with a bit/word trade-off that stays predictable by design. Its weaknesses concentrate where the pretrained denoiser is mismatched to the attack, i.e., -- strong filtering, restoration, and noise, -- rather than in the synchronization or coding stages.

Because the design is synchronization-based, modular, and not learned end to end, each stage can be improved in isolation without retraining the others, so the same construction can be pushed further toward any given attack, transmission channel, or payload regime. We see three directions. First, replacing or fine-tuning the host predictor for extraction under filtering, blur, and noise directly targets the main limitation while leaving the rest of the pipeline untouched. Second, extending the acquisition model from affine to general projective warps would close the remaining geometric gap. Third, since the carrier expands by plain repetition, its scale, code rate, and synchronization budget can be retuned per channel for a specific compression, display, or distribution channel trading capacity for robustness on demand.

\clearpage
\balance
\bibliographystyle{IEEEbib}
\begingroup
\interlinepenalty=10000
\bibliography{references}
\endgroup
\end{document}

%% file: figures/pipeline.tex
\begin{figure*}[!t]
\vspace{-10mm}
\centering
\begin{tikzpicture}[x=1mm,y=1mm,>=Latex,
box/.style={draw=black!60,fill=blue!7,align=center,text width=18mm,
minimum height=12mm,font=\fontsize{7.6}{8.4}\selectfont,inner sep=1.8pt},
evidence/.style={box,fill=green!9},
learned/.style={box,fill=orange!17},arr/.style={->,semithick}]
\node[evidence] (msg) at (10,0) {$L_{\rm msg}$ message\\bits; append\\$L_{\rm CRC}$ bits};
\node[box] (code) at (31,0) {Encrypt word\\Polar $(N,L)$};
\node[evidence] (layout) at (52,0) {Keyed sync $\mathbf s$\\layout $(\mathbf c,\mathbf s)$};
\node[box] (bipolar) at (73,0) {Bipolar map\\$0\!\mapsto\!+1$\\$1\!\mapsto\!-1$};
\node[box] (tile) at (73,-25) {Expand cells\\and tile $\mathbf w$};
\node[box] (mask) at (52,-25) {Perceptual mask\\$\alpha\mathbf m\odot\mathbf w$};
\node[box] (add) at (31,-25) {Add to host\\clip; quantize};
\node[evidence] (y) at (10,-25) {Marked image\\$\mathbf y$};
\node[draw=black!60,fill=green!9,rounded corners=1pt,
font=\fontsize{7.6}{8.4}\selectfont,inner sep=2pt] (host) at (41.5,-12.5) {Host image $\mathbf x$};
\node[draw=black!60,fill=white,rounded corners=1pt,
font=\fontsize{7.6}{8.4}\selectfont,inner sep=2pt] (keyE) at (41.5,12) {$k_p,\nu$};
\draw[arr](keyE.south west)--(code.north);
\draw[arr](keyE.south east)--(layout.north);
\draw[arr](msg)--(code);\draw[arr](code)--(layout);\draw[arr](layout)--(bipolar);
\draw[arr](bipolar)--(tile);\draw[arr](tile)--(mask);\draw[arr](mask)--(add);\draw[arr](add)--(y);
\draw[arr](host.south west)--(add.north);
\draw[arr](host.south east)--(mask.north);
\node[font=\fontsize{9}{10}\selectfont] at (41.5,-35) {(a) Embedding.};
\node[evidence] (z) at (102,0) {Received image\\$\mathbf z$};
\node[learned] (pred) at (123,0) {Neural host\\suppression\\$\widehat{\mathbf w}=\mathbf z-D(\mathbf z)$};
\node[box] (acf) at (144,0) {Residual ACF\\+ RANSAC};
\node[evidence] (affine) at (165,0) {Affine matrix\\hypotheses $\mathbf A$};
\node[box] (fold) at (165,-25) {Inverse warp\\fold; average};
\node[box] (sync) at (144,-25) {Orient; correlate\\keyed sync cells\\translation/phase};
\node[box] (decode) at (123,-25) {Polar decode\\decrypt\\CRC equality};
\node[evidence] (result) at (102,-25) {Recovered bits\\or failure};
\node[draw=black!60,fill=white,rounded corners=1pt,
font=\fontsize{7.6}{8.4}\selectfont,inner sep=2pt] (keyD) at (133.5,-12.5) {$k_p,\nu$};
\draw[arr](keyD.south west)--(decode.north);
\draw[arr](keyD.south east)--(sync.north);
\draw[arr](z)--(pred);\draw[arr](pred)--(acf);\draw[arr](acf)--(affine);
\draw[arr](affine)--(fold);\draw[arr](fold)--(sync);\draw[arr](sync)--(decode);\draw[arr](decode)--(result);
\node[font=\fontsize{9}{10}\selectfont] at (133.5,-35) {(b) Blind extraction and CRC check.};
\end{tikzpicture}
\vspace{-4mm}
\caption{NAWE signal path. Blue blocks perform analytical processing, green blocks carry image or message evidence, and orange denotes the frozen neural host predictor. The host determines the perceptual mask; the independent message and keyed synchronization bits determine the carrier. The receiver shares the key, nonce, layout, and decoder settings.}
\label{fig:pipeline}
\end{figure*}